\documentclass{article}

\usepackage{graphicx} 
\usepackage[usenames,dvipsnames]{xcolor}
\usepackage{subfigure} 
\usepackage{amsmath, amsthm, amssymb, bm, bbm}
\usepackage{natbib}
\usepackage{mathrsfs}
\usepackage{bbm}
\usepackage{multirow}
\usepackage{array}
\newcolumntype{C}[1]{>{\centering\let\newline\\\arraybackslash\hspace{0pt}}m{#1}}
\newcolumntype{L}[1]{>{\raggedright\let\newline\\\arraybackslash\hspace{0pt}}m{#1}}
\newcolumntype{R}[1]{>{\raggedleft\let\newline\\\arraybackslash\hspace{0pt}}m{#1}}
\usepackage{algorithm}
\usepackage{algorithmic}
\usepackage{hyperref}

\usepackage{tikz}
\usepackage{pgfplots}
\usepackage{wasysym}

\newcommand*{\vpointer}{\vcenter{\hbox{\scalebox{2}{$\xrightarrow{\text{shifted}}$}}}}
\pgfplotsset{compat=newest} 
\pgfplotsset{plot coordinates/math parser=false}
\usetikzlibrary{plotmarks}

\newcommand\numberthis{\addtocounter{equation}{1}\tag{\theequation}}
\usepackage[accepted]{icml2013stylefiles/icml2013}

\def\IBP{\boldsymbol{Z}}
\def\Features{\boldsymbol{A}}
\def\feat{a}
\def\Data{\boldsymbol{X}}
\def\datum{x}
\def\ibp{z}

\def\OptMat{\boldsymbol{\omega}}

\def\Errors{\boldsymbol{E}}
\def\Hypers{\boldsymbol{\theta}}

\def\sigData{\sigma_{\Data}}
\def\sigFeat{\sigma_{\Features}}

\def\TN{T\mathcal{N}}

\def\ExpFM{\boldsymbol{\Phi}}
\def\lcdot#1{_{#1\cdot}}
\def\erfcx#1{\text{erfcx}\left(#1\right)}
\DeclareMathOperator*{\argmax}{arg\,max}
\DeclareMathOperator*{\E}{\mathbb{E}}
\newtheorem{proposition}{Proposition}

\icmltitlerunning{Scaling the Indian Buffet Process via Submodular Maximization}

\begin{document} 
\twocolumn[
\icmltitle{Scaling the Indian Buffet Process via Submodular Maximization}

\icmlauthor{Colorado Reed}{cr478@cam.ac.uk}
\icmlauthor{Zoubin Ghahramani}{zoubin@eng.cam.ac.uk}
\icmladdress{Engineering Department, Cambridge University, Cambridge UK}

\icmlkeywords{indian buffet process, submodular optimization, machine learning, ICML}

\vskip 0.3in
]

\begin{abstract} 
Inference for latent feature models is inherently difficult as the inference space grows exponentially with the size of the input data and number of latent features. In this work, we use \citet{kurihara2008bayesian}'s maximization-expectation framework to perform approximate MAP inference for linear-Gaussian latent feature models with an Indian Buffet Process (IBP) prior. This formulation yields a submodular function of the features that corresponds to a lower bound on the model evidence. By adding a constant to this function, we obtain a nonnegative submodular function that can be maximized via a greedy algorithm that obtains at least a $\frac{1}{3}$-approximation to the optimal solution. Our inference method scales linearly with the size of the input data, and we show the efficacy of our method on the largest datasets currently analyzed using an IBP model. 
\end{abstract} 

\section{Introduction}
Nonparametric latent feature models experienced a surge of interest in the machine learning community following \citet{griffiths2006infinite}'s formulation of the Indian Buffet Process (IBP)---a nonparametric prior for equivalence classes of sparse binary matrices. These binary matrices have a finite number of exchangeable rows and an unbounded number of columns, where a 1 in row $n$ and column $k$ indicates that observation $n$ expresses latent feature $k$. For example, given an image dataset of human faces, each observation is an image, and the latent features might be ``is smiling," ``is wearing glasses," etc. More generally, feature models can be viewed as a generalization of unsupervised clustering, see \citet{broderick2012clusters}.

The IBP prior is often used in sparse matrix factorization models where a data matrix of $N$ $D$-dimensional observations is expressed as a product of two matrices that factor over $K$ latent factors plus a noise term: $\Data = \IBP\Features + \Errors$. Formally, this model has a binary feature matrix $\IBP \in \{0,1\}^{N \times K}$ that linearly combines a latent factor matrix $\Features \in \mathbb{R}^{K \times D}$ plus a noise matrix $\Errors \in \mathbb{R}^{N\times D}$ to form the observed data $\Data \in \mathbb{R}^{N\times D}$. Placing an IBP prior on $\IBP$ lets $K$ be unbounded and allows the number of active features $K_+$ (those with non-zero $\IBP$ column sums) to be learned from the data while remaining finite with probability one. The IBP inspired several infinite-limit versions of classic matrix factorization models, e.g.\ infinite independent component analysis models \citep{knowles2007infinite}.

Inference with IBP models is challenging as its discrete state space has $2^{NK_+}$ possible assignments. In turn, the IBP has found limited application to large data in comparison to the Chinese Restaurant Process, which assigns one feature to each observation.
In this paper, we use \citet{kurihara2008bayesian}'s Maximization-Expectation (ME) framework to perform approximate MAP inference with IBP matrix factorization models, termed MEIBP inference. For nonnegative $\Features$, we show that we can obtain approximate MAP solutions for $\IBP$ by maximizing $N$ submodular cost functions. The submodularity property enables the use of a simple greedy algorithm that obtains at least a $\frac{1}{3}$-approximation to the optimal solution. While the worst-case complexity of MEIBP inference is comparable to sampling and variational approaches, in $\S\ref{sec:exps}$ we show that MEIBP inference often converges to better solutions than variational methods and similar solutions as the best sampling techniques but in a fraction of the time.

This paper is structured as follows: in $\S$\ref{sec:background} we present background material that sets the foundation for our presentation of MEIBP inference in $\S$\ref{sec:meibp} and the resulting submodular maximization problem that arises, then  in $\S$\ref{sec:relwork} we discuss related work, and in $\S$\ref{sec:exps} we compare the MEIBP with other IBP inference techniques using both synthetic and real-world datasets.

\section{Background}\label{sec:background}
\subsection{The Indian Buffet Process}
\citet{griffiths2006infinite} derived the IBP prior by placing independent beta priors on Bernoulli generated entries of an $N \times K$ binary matrix $\IBP$, marginalizing over the beta priors, and letting $K$ go to inifinity. In this infinite limit, however, $P(\IBP)$ is zero for any particular $\IBP$. \citet{griffiths2006infinite} therefore take the limit of an equivalence classes of binary matrices, $[\IBP]$, defined by the ``left-order form" (\textit{lof}) ordering of the columns and show that $P([\IBP]_{\textit{lof}})$ has a non-zero probability as $K$ goes to infinity.

The \textit{lof} ordering arranges the columns of $\IBP$ such that the binary values of the columns are non-increasing, where the first row is the most significant bit. \citet{ding2010nonparametric} examine different ``shifted" equivalence classes formed by shifting all-zero columns to the right of non-zero columns while maintaining the non-zero column ordering. Given $K_+$ non-zero columns, the IBP prior for the shifted equivalence classes is
\begin{align}\label{ibpprior}
P([\IBP]|\alpha) = \frac{\alpha^{K_+}}{K_+!}\text{e}^{-\alpha H_N} \prod_{k=1}^{K^{+}} \frac{(N-m_k)!(m_k - 1)!}{N!}
\end{align}
where $\alpha$ is a hyperparameter, $H_N$ is the $N^{th}$ harmonic number, and $m_k = \sum_{n=1}^N \ibp_{nk}$.
The supplementary material has a derivation of $P([\IBP]|\alpha)$ as well as a comparison to the \textit{lof} equivalence classes. The derivations in $\S$\ref{sec:meibp} can be applied using either equivalence class. However, the shifted equivalence classes simplify the mathematics and produce the same results in practice.

\subsection{Maximization-Expectation}
\citet{kurihara2008bayesian} presented the ME algorithm: an inference algorithm that exchanges the expectation and maximization variables in the EM algorithm. Consider a general probabilistic model $p(\Data, \IBP, \Features)$, where $\Data$ are the observed random variables (RVs), $\IBP$ are the local latent RVs, and $\Features$ are the global latent RVs. RVs are qualified as ``local" if there is one RV for each observation, and RVs are ``global" if their multiplicity is constant or inferred from the data.

ME can be viewed as a special case of a Mean-Field Variational Bayes (MFVB) approximation to a posterior that cannot be computed analytically, $p(\IBP, \Features|\Data)$. MFVB operates by approximating the posterior distribution of a given probabilistic model by assuming independent variational distributions, $p(\IBP, \Features|\Data) \approx q(\IBP)q(\Features)$ \citep{attias2000variational,ghahramani2001propagation}. The independence constraint lets us compute the variational distribution $q$ that minimizes the KL divergence between the variational distribution and true posterior. Without this constraint, the distribution that minimizes the KL-divergence is the true posterior, returning us to our original problem. In MFVB, we determine the variational distributions and their parameters using coordinate ascent optimization in which we iteratively update:
\begin{align}
q(\IBP) &\propto  \text{exp} \E_{q(\Features)}\left[\ln{p(\Data, \IBP, \Features)}\right]\\
q(\Features) &\propto  \text{exp}\E_{q(\IBP)}\left[\ln{p(\Data, \IBP, \Features)}\right],
\end{align}
which commonly has closed-form solutions.

The EM algorithm can be viewed as a special case of MFVB that obtains MAP values of the global RVs by letting $q(\Features) = \delta(\Features - \Features^{*})$, where $\delta(\cdot)$ is the delta function and $\Features^{*}$ is the MAP assignment. The ME algorithm instead maximizes the local RVs $\IBP$ and computes the expectation over the global RVs $\Features$, 
which can be viewed as MFVB with $q(\IBP) = \delta(\IBP - \IBP^{*})$. In the limit of large $N$, the ME algorithm recovers a Bayesian information criterion regularization term \citep{kurihara2008bayesian}.  Also, maintaining a variational distribution over the global RVs retains the model selection ability of MFVB, while using point estimates of the local RVs allows the use of efficient data structures and optimization techniques. As we will show, the ME algorithm leads to a scalable submodular optimization problem for latent feature models.

\subsection{Submodularity}\label{sec:submodopt}
Submodularity is a set function property that makes optimization of the function tractable or approximable. Given ground set $V$ and set function $f:2^V \rightarrow \mathbb{R}$, $f$ is \textit{submodular} if for all $A \subseteq B \subseteq V$ and $e \in V \backslash B$:
\begin{align} \label{eq:submoddef}	
 f(A \cup \{e\}) - f(A) \geq  f(B \cup \{e\}) - f(B),
\end{align}
which expresses a ``diminishing returns" property, where the incremental benefit of element $e$ diminishes as we include it in larger solution sets. Submodularity is desirable in discrete optimization because submodular functions are discrete analogs of convex functions and can be globally minimized in polynomial time \citep{lovasz1983submodular}.  However, global submodular maximization is NP-hard, but submodularity often enables approximation bounds via greedy algorithms.  In the next section, we show that determining a MAP estimate of $\IBP$ in the ME algorithm is a scalable submodular maximization problem.

\section{Maximization-Expectation IBP}\label{sec:meibp}
Here we present the ME algorithm for nonnegative linear-Gaussian IBP models and show that approximate MAP inference arises as a submodular maximization problem. Boldface variables are matrices with (row, column) subscripts; a dot indicates all elements of the dimension, and lowercase variables are scalars.

\subsection{Nonnegative Linear-Gaussian IBP Model}
We consider the following probabilistic model:
\begin{align}\label{jointp}
p(\Data, \IBP, \Features|\Hypers) &= p(\Data|\IBP, \Features, \sigData^2) p(\Features|\sigFeat^2) p(\IBP|\alpha) \\
p(\Data|\IBP, \Features, \sigFeat^2) &=  \prod_{n=1}^N \mathcal{N}(\Data\lcdot{n}; \IBP\lcdot{i}\Features, \sigFeat^2 I) \\
p(\Features|0, \sigFeat^2) &= \prod_{k=1}^K\prod_{d=1}^D \TN(\feat_{kd}; 0,\sigFeat^2)
\end{align}
with $p([\IBP]|\alpha)$ specified in Eq.\ \ref{ibpprior}. This is a nonnegative linear-Gaussian IBP model, where the prior over the latent factors, $p(\Features|0, \sigFeat^2)$, is a zero-mean i.i.d.\ truncated Gaussian with nonnegative support, denoted $\TN$.  As we show below, this nonnegative prior yields a submodular maximization problem when optimizing $\IBP$.  We use a truncated Gaussian as it is conjugate to the Gaussian likelihood, but other nonnegative priors (e.g.\ exponential) can be used. For brevity we assume the hyperparameters, $\Hypers = \{\alpha,\sigFeat^2,\sigData^2\}$, are known and discuss $\Hypers$ inference in the supplementary material.

\subsection{MEIBP Evidence}
In the ME framework, we approximate the true posterior distribution via a MFVB assumption:
\begin{equation}\label{varfactor}
p(\IBP, \Features|\Data,\Hypers) \approx q(\Features)\delta(\IBP - \IBP^{*}).
\end{equation}
That is, we maintain a variational distribution over the latent factors $\Features$ and optimize the latent features $\IBP$. Given the MFVB constraint, we determine the variational distributions by minimizing the KL-divergence between the variational distributions and the true posterior, which is equivalent to maximizing a lower bound on the evidence \citep{attias2000variational}:
 \begin{align*}
 \ln p(\Data|\Hypers) & = \E_q[\ln p(\Data,\Features,\IBP|\Hypers)] + H[q] + D(q\|p)\\
  & \geq \E_q[\ln p(\Data,\Features,\IBP|\Hypers)] + H[q] \equiv \mathcal{F}\label{eveq} \numberthis
 \end{align*}
where $H[q]$ is the entropy of $q$ and $D(q\|p)$ represents the KL-divergence between the variational distribution and the true posterior. 
The evidence lower bound, $\mathcal{F}$, for the nonnegative linear-Gaussian IBP model is:
\begin{align}
&\frac{1}{\sigData^2}\sum_{n=1}^N\left[ -\frac{1}{2}\IBP\lcdot{n}\ExpFM\ExpFM^{T}\IBP\lcdot{n}^{T} + \IBP\lcdot{n}\boldsymbol{\xi}\lcdot{n}^T\right] - \ln{K_+!} \nonumber \\
 &~+ \sum_{k=1}^{K^+}\left[\ln{\frac{(N-m_k)!(m_k - 1)!}{N!}}
 + \eta_k \right]
+  \text{const} \label{vlb}
\end{align}
with
\begin{align}
&\xi_{nk} = \ExpFM\lcdot{k}\Data\lcdot{n}^T + \frac{1}{2}\sum_{d=1}^D\left[\E[\feat_{kd}]^2 - \E[\feat_{kd}^2] \right] \label{xiterm}
\end{align}
and
\begin{align}
&\eta_k = \sum_{d=1}^D\Bigl[ - \frac{\ln{\frac{\pi \sigFeat^2}{2\alpha^{2/D}}}}{2} - \frac{\E[a_{kd}^2]}{2\sigFeat^2} + H(q(\feat_{kd})) \Bigr]
\end{align}
where $\ExpFM\lcdot{k} = \left(\E\left[ a_{k1}\right], \ldots,  \E\left[ a_{kD}\right]  \right)$, and all expectations are with respect to $q(\Features)$, which is defined in the next subsection. In $\S$\ref{sec:zobj} we show that maximizing this lower bound with respect to $\IBP$ can be formulated as a submodular maximization problem.

\subsection{Variational Factor Updates}
Maximizing Eq.\ \ref{vlb} with respect to $q(\Features)$ yields 
\begin{align}
q(\Features) & = \prod_{k=1}^K \prod_{d=1}^D \TN(\feat_{kd};\tilde{\mu}_{kd}, \tilde{\sigma}^2_{kd}),
\end{align} 
with parameter updates
\begin{align}
\tilde{\mu}_{kd} &=   \rho_k
		 \sum_{n=1}^N \ibp_{nk} \Bigl( \datum_{nd}  - \sum_{k'\neq k} \ibp_{nk'} \E\left[ \feat_{k'd} \right] \Bigr) \\
\tilde{\sigma}^2_{kd} &= \rho_k \sigma^2_X ,
\end{align}
where $\rho_k = \bigl(m_{k} + \frac{\sigma_X^2}{\sigFeat^2}\bigr)^{-1}$. These updates take $O(NK^2D)$, and the relevant moments are:
\begin{align}
\mathbb{E}\left[ \feat_{kd} \right] &= \tilde{\mu}_{kd} 
	+  \tilde{\sigma}_{kd}
	\frac{\sqrt{2/\pi}}{\erfcx{\wp_{kd}}}  \label{exa}
	\\
\mathbb{E}\left[ \feat_{kd}^2 \right] &= \tilde{\mu}_{kd}^2 + \tilde{\sigma}_{kd}^2 
	+   \tilde{\sigma}_{kd} \tilde{\mu}_{kd} 
	\frac{\sqrt{2/\pi}}{\erfcx{\wp_{kd}}} \label{exasq}
\end{align}
with $\wp_{kd}=-\frac{\tilde{\mu}_{kd} }{\tilde{\sigma}_{kd}\sqrt{2}}$ and $\erfcx{y}=e^{y^2}(1-\text{erf}(y))$ representing the scaled complementary error function.

\subsection{Evidence Lower Bound as $K\rightarrow \infty$}
Here we show that the evidence lower bound [Eq.\ \ref{vlb}] is well-defined in the limit $K \rightarrow \infty$; in fact, all instances of $K$ are simply replaced by $K_+$. Therefore, similar to variational IBP methods, a user must specify a maximum model complexity $K_+$. A benefit over variational IBP methods, however, is that the $q(\IBP)$ updates are not affected by inactive features|see $\S\ref{sec:relwork}$. 

We take this limit by breaking the evidence into components $1,\ldots, K_{+}$ and $K_{+} + 1, \ldots, K$ and note that when $m_k = 0$: $\tilde{\mu}_{kd} =  0$, $\tilde{\sigma}^2_{kd} = \sigFeat^2$, and $H(a_{kd})=\frac{1}{2}\ln{\frac{\pi e \sigFeat^2}{2}}$. After some algebra, the evidence becomes:
\begin{align} \label{inactlb}
\psi_{K_+} + \frac{1}{2}\sum_{k=K_++1}^K \sum_{d=1}^D\left[ -\ln{\frac{\pi \sigFeat^2}{2}} - \frac{\E[\feat_{kd}^2]}{\sigFeat^2} + \ln{\frac{\pi e \sigFeat^2}{2}} \right]
\end{align} 

where $\psi_{K_+}$ is Eq.\ \ref{vlb} but with $K_+$ replacing all $K$. From Eq.\ \ref{exasq}, we see that $\E[\feat_{kd}^2]= \sigFeat^2$ when $m_k=0$, which causes all terms to cancel in Eq.\ \ref{inactlb} except $\psi_{K_+}$.

The evidence lower bound remains well-defined because both the likelihood and IBP prior terms do not depend on inactive features, so for inactive features the KL-divergence between the posterior and variational distributions is simply the KL-divergence between $p(\Features)$ and $q(\Features)$. For inactive features, $p(\Features) = q(\Features)$, and as a result, the KL-divergence is zero.

\subsection{$\IBP$ Objective Function}\label{sec:zobj}
Given $q(\Features)$, we compute MAP estimates of $\IBP$ by maximizing the evidence [Eq.\ \ref{vlb}] for each $n\in\{1,\ldots,N\}$ while holding constant all $n'\in \{1,\ldots,N\}\setminus n$. Decomposing Eq.\ \ref{vlb} into terms that depend on $\IBP\lcdot{n}$ and those that do not yields (see the supplementary material):
\begin{align}
 \mathcal{F}(\IBP\lcdot{n}) =& -\frac{1}{2\sigData^2}\IBP\lcdot{n}\ExpFM\ExpFM^{T}\IBP\lcdot{n}^{T} + \IBP\lcdot{n}\OptMat\lcdot{n}^{T} + \textit{const} \nonumber\\
  &-  \ln{ \Bigl(K_{+\setminus n} + \sum_{k=1}^{K_+}\left[ \boldsymbol{1}_{\{m_{k\setminus n} = 0\}}\ibp_{nk}\right] \Bigr)!} \label{subcf} \\
\ExpFM\lcdot{k} =& \Bigl(\E\left[ a_{k1}\right], \ldots,  \E\left[ a_{kD}\right]  \Bigr) \nonumber \\
\omega_{nk} =& \frac{1}{\sigData^2}\Bigl(\ExpFM\lcdot{k}\Data\lcdot{n}^T + \frac{1}{2}\sum_{d=1}^D\left[\E[\feat_{kd}]^2 - \E[\feat_{kd}^2] \right]\Bigr) \nonumber \\
& +  \nu(\ibp_{nk}=1) - \nu(\ibp_{nk}=0)+ \boldsymbol{1}_{\{m_{k\setminus n} = 0\}}\eta_k  \nonumber,Ê
\end{align}
which is a quadratic pseudo-Boolean function plus a term that penalizes $K_{+}$,  where $\boldsymbol{1}_{\{ \cdot \}}$ is the indicator function, a ``$\setminus n$" subscript indicates the given variable is determined after removing the $n^{\text{th}}$ row from $\IBP$, and 
\begin{align*} 
  \nu(\ibp_{nk}) = \begin{cases}
    0,  ~~\text{if $m_{k\setminus n} = 0$ and $\ibp_{nk} = 0$}\\
    \ln{(N-m_{k\setminus n} - \ibp_{nk})!/N!}\\ 
    ~~+\ln{(m_{k \setminus n} + \ibp_{nk} - 1)!}, ~~\text{otherwise}  \label{nueq}\\
  \end{cases}
\end{align*}
 We can prove $\mathcal{F}(\IBP\lcdot{n})$ is submodular given the following two well-known propositions, see \citet{fujishige2005submodular}:
\begin{proposition} \label{nnlc_prop}
Nonnegative linear combinations of submodular functions are submodular.
\end{proposition}
\begin{proposition}\label{qpbs_prop}
A quadratic pseudo-Boolean function with quadratic weight matrix $\mathbf{W}$  is submodular if and only if $W_{ij} \leq 0$ for all $i,j$. 
\end{proposition}
Via Proposition \ref{qpbs_prop}, we see that $-\frac{1}{2\sigData^2}\IBP\lcdot{n}\ExpFM\ExpFM^{T}\IBP\lcdot{n}^{T} + \IBP\lcdot{n}\OptMat\lcdot{n}^{T}$ is submodular when $\ExpFM$ is nonnegative. From Proposition \ref{nnlc_prop}, Eq.\ \ref{subcf} is submodular if and only if 
\begin{align}
\mathcal{G}(\IBP\lcdot{n}) = - \ln{ \Bigl(K_{+\setminus n} + \sum_{k=1}^{K_+}\left[ \boldsymbol{1}_{\{m_{k\setminus n} = 0\}}\ibp_{nk}\right] \Bigr)!}
\end{align}
 is submodular. We prove this property by rephrasing $\mathcal{G}(\IBP\lcdot{n})$ as a set function and using the definition of submodularity given by Eq.\ \ref{eq:submoddef}. Let $A_n \subseteq B_n \subseteq V$ where $V=\{1, \ldots, K_{+}\}$ and $A_n,B_n \in 2^{V}$ with 
$\mathcal{G}(A_n) = - \ln{(K_{+\setminus n} + K_{A_n})!}$.
Here we let $K_{A_n} = \sum_{k=1}^{K_+}\boldsymbol{1}_{\{m_{k\setminus n} = 0\}} \boldsymbol{1}_{\{ k \in A_n\}}$ where $k \in A_n$ indicates $\ibp_{nk}=1$. $\mathcal{G}$ is submodular if for all $e \in V\setminus B_n$:
\begin{align*}
\mathcal{G}(A_n \cup \{e\}) - \mathcal{G}(A_n) & \geq \mathcal{G}(B_n \cup \{e\}) - \mathcal{G}(B_n)\\
\ln{\frac{\left(K_{+\setminus n} +  K_{A_n}\right)!}{\left(K_{+\setminus n} +  K_{A_n \cup \{e\}}\right)!}} &\geq \ln{\frac{\left(K_{+\setminus n} +  K_{B_n}\right)!}{\left(K_{+\setminus n} +  K_{B_n \cup \{e\}}\right)!}} \label{eq:submodineq}\numberthis
\end{align*}
Eq.\ \ref{eq:submodineq} has two cases: (1) $m_{e\setminus n} > 0$ so $K_{B_n \cup \{e\}}=K_{B_n}$ and $K_{A_n \cup \{e\}}=K_{A_n}$, yielding $0 \geq 0$ for Eq.\ \ref{eq:submodineq}, which is true for all $e \in V\setminus B_n$ and $A_n\subseteq B_n$, (2) $m_{e\setminus n} = 0$ so $K_{B_n \cup \{e\}}=K_{B_n}+1$ and $K_{A_n \cup \{e\}}=K_{A_n}+1$.  After some algebra this yields $K_{B_n\cup\{e\}} \geq K_{A_n\cup \{e\}}$ for Eq.\ \ref{eq:submodineq}, which is again true for all $e \in V\setminus B_n$ and $A_n\subseteq B_n$. As a result, both components of Eq.\ \ref{subcf} are submodular, and by Proposition \ref{nnlc_prop}, adding these  terms yields a submodular function.

\subsection{$\IBP$ Optimization}\label{sec:zopt}
Eq.\ \ref{subcf} is an unconstrained nonmonotone submodular function. \citet{feige2011maximizing} prove that an approximibility guarantee is NP-hard for this class of functions. However, \citet{feige2011maximizing} also show that a local-search (ls) algorithm obtains a constant-factor approximation to the optimal solution, provided the submodular objective function is nonnegative. For a submodular function $\mathcal{F}:2^{V}\rightarrow \mathbb{R}$ with ground set $V=\{1,\ldots,K_+\}$ and solution set $A\subseteq V$, the ls-algorithm operates as follows:
\begin{enumerate}
\item \textit{initialize}: let $A=\{\argmax_{w\in V}\mathcal{F}(\{w\})\}$
\item \textit{grow}: while there is an element $w \in V \setminus A$ s.t.\ $\mathcal{F}(A \cup \{w\}) > (1+\frac{\epsilon}{|V|^2})\mathcal{F}(A)$:  let $A := A\cup\{w\}$
\item \textit{prune}: if there is an element $w \in A$ s.t.\ $\mathcal{F}(A\setminus\{w\}) > (1+\frac{\epsilon}{|V|^2})\mathcal{F}(A)$: let $A := A\setminus\{w\}$, goto 2.
\item \textit{return}: maximum of $\mathcal{F}(A)$ and $\mathcal{F}(V\setminus A)$.
\end{enumerate}
The ls-algorithm obtains a solution that is greater than $\frac{1}{3}(1 - \frac{\epsilon}{|V|})\text{OPT}$|where $\epsilon$ is a parameter and $\text{OPT}$ is the maximum value of $\mathcal{F}$. The ls-algorithm performs $O(\frac{1}{\epsilon}|V|^3\log{|V|})$ function calls in the grow/prune steps. 

\begin{figure}[htb]
	\centering
	\newlength\fheight 
	\newlength\fwidth 
	\setlength\fheight{2.9cm} 
	\setlength\fwidth{7cm}
%
%
%
%
\begin{tikzpicture}

\begin{axis}[%
width=\fwidth,
height=\fheight,
scale only axis,
xmin=1, xmax=13,
xtick={2,4,6,8,10,12},
xlabel={\footnotesize latent features (K)},
ymin=0, ymax=1.05,
ytick={0,0.2,0.4,0.6,0.8,1},
ylabel={\footnotesize fraction of opts},
ymajorgrids,
yminorgrids,
minor ytick={0.1,0.3,...,0.9},
legend style={at={(0.277940345846368,0.14250885857181)},anchor=south west,draw=black,fill=white,align=left,font=\scriptsize,nodes={right}}]

\addplot [
color=gray!60!black,
solid,
line width=1.0pt,
mark size=1.5pt,
mark=o,
mark options={solid,draw=darkgray!80!black}
]
plot [error bars/.cd, y dir = both, y explicit]
coordinates{
 (2,1)+-(0.0,0)(4,0.99685)+-(0.0,0.00241994031147696)(6,0.9978743)+-(0.0,0.00368824291348735)(8,0.9989738)+-(0.0,0.000840865797192891)(10,0.9987514)+-(0.0,0.00331634558579841)(12,0.9998572)+-(0.0,0.000308503844744568) 
};
\addlegendentry{$\text{LS }\geq\text{ 95\% Optimum}$};

\addplot [
color=gray!60!black,
solid,
line width=1.0pt,
mark size=1.5pt,
mark=x,
mark options={solid,draw=darkgray!80!black}
]
plot [error bars/.cd, y dir = both, y explicit]
coordinates{
 (2,1)+-(0.0,0)(4,0.9843733)+-(0.0,0.00714814163961519)(6,0.9469089)+-(0.0,0.0167191661969794)(8,0.8834671)+-(0.0,0.0212504066546712)(10,0.8053402)+-(0.0,0.0377291720962264)(12,0.7254797)+-(0.0,0.0432624780100108) 
};
\addlegendentry{LS Finds Optimum};

\addplot [
color=gray!60!black,
dashed,
line width=1.0pt,
mark size=1.5pt,
mark=o,
mark options={solid,draw=darkgray!80!black}
]
plot [error bars/.cd, y dir = both, y explicit]
coordinates{
 (2,0.266)+-(0.0,0.00892759351673228)(4,0.1043767)+-(0.0,0.00771774189845133)(6,0.06448308)+-(0.0,0.00687704923589244)(8,0.04934944)+-(0.0,0.00642466818065771)(10,0.04770148)+-(0.0,0.00479565272832246)(12,0.04707968)+-(0.0,0.0075806232673691) 
};
\addlegendentry{$\text{Random  }\geq\text{  95\% Optimum}$};

\addplot [
color=gray!60!black,
dashed,
line width=1.0pt,
mark size=1.5pt,
mark=x,
mark options={solid,draw=darkgray!80!black}
]
plot [error bars/.cd, y dir = both, y explicit]
coordinates{
 (2,0.2489333)+-(0.0,0.00764768471935567)(4,0.06558667)+-(0.0,0.00399966765287373)(6,0.01555738)+-(0.0,0.00219355588273166)(8,0.003335794)+-(0.0,0.00104619752605115)(10,0.0008023302)+-(0.0,0.000340117638597719)(12,0.0002722223)+-(0.0,0.000237684988729971) 
};
\addlegendentry{Random Finds Optimum};

\end{axis}
\end{tikzpicture}%
	\caption{Fraction of ls-algorithm and random solutions that obtain [within $95\%$ of] the true optimum using data generated from the nonnegative linear-Gaussian model with $N=500, D=50, \sigma_X = 1.0$.}
	\label{fig:lscomp}
\end{figure}
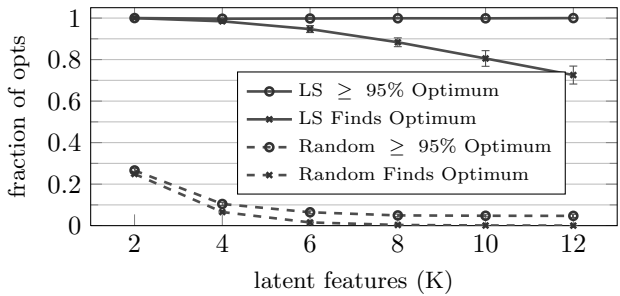

Since Eq.\ \ref{subcf} is not strictly nonnegative, we use its \textit{normalized cost function} to interpret the ls-approximability guarantee: $\mathcal{F}(\IBP\lcdot{n}) - \mathcal{F}_{n0}$, where $\mathcal{F}_{n0}$ is the minimum value of $\mathcal{F}(\IBP\lcdot{n})$. Using the normalized cost function, we obtain the following optimality guarantee:
\begin{align}\label{submod_lb}
\mathcal{F}(\IBP\lcdot{n}^{\text{ls}}) \geq \mathcal{F}_{n0}+ \frac{1}{3}\left(1-\frac{\epsilon}{|V|}\right)\left(\mathcal{F}(\IBP\lcdot{n}^{*}) - \mathcal{F}_{n0}\right)
\end{align}
where the superscript ``ls" denotes the solution from the greedy ls-algorithm and an asterisk denotes the set that obtains the true maximum. This inequality states that the ls-algorithm solution is guaranteed to perform better than the minimum by an amount proportional to the difference between the optimum and the minimum. However, we emphasize that this inequality does not provide an optimality guarantee for the global MAP solution.

We studied the empirical performance of the ls-algorithm by generating high noise data ($\sigma_X = 1$) from the nonnegative linear-Gaussian model with $N=500, D=50$ and compared the ls-algorithm with the brute-force optimal solution as $K$ varied from $2$ to $12$, performing $1000K$ total optimizations for each of ten randomly generated datasets. Furthermore, we compared the ls-algorithm with randomly sampled $\IBP\lcdot{n}$ solutions to demonstrate that the optimization space was not skewed to favor solutions near the optimal value. 

Figure \ref{fig:lscomp} shows the fraction of solutions that obtain the true optimum as well as the fraction of solutions that were greater than $95\%$ of $\mathcal{F}(\IBP\lcdot{n}^{*}) - \mathcal{F}_{n0}$, where the error bars indicate the combined standard deviation over the $10\times1000K$ optimizations. The ls-algorithm found the optimal solution roughly $70\%$ of the time for $K=12$ and obtained within $95\%$ of the optimal solution over $99.9\%$ of the time for all $K$|meaning we could empirically replace the $\frac{1}{3}$ in Eq.\ \ref{submod_lb} with $\frac{19}{20}$. The random sampling comparison indicated that the optimization space did not favor nearly-optimal solutions: its convergence to $5\%$ for \textit{within}-$95\%$ \textit{optimal solutions} was characteristic of a uniform solution space.

\begin{figure}[thb]
	\centering
	\setlength\fheight{2.2cm} 
	\setlength\fwidth{6.7cm}
	\input{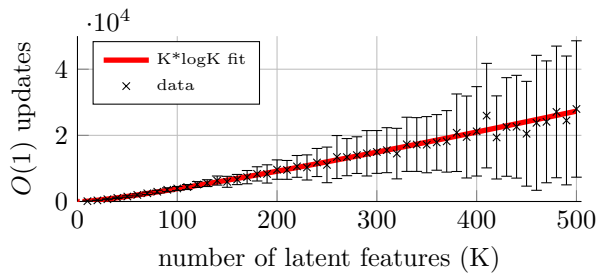}
	\caption{Number of $O(1)$ updates per ls-optimization using data generated from the nonnegative linear-Gaussian model with $N=1000, D=1000, \sigma_X = 1.0$. }
	\label{fig:rtcomp}
\end{figure}

By precomputing  $\ExpFM\ExpFM^{T}$ and maintaining an auxiliary vector of $K_+$  weights, we can evaluate Eq.\ \ref{subcf} in constant time when adding/removing elements to the solution set. In turn, the ls-algorithm optimizes $\mathcal{F}(\IBP\lcdot{n})$ in $K_+^2D + O(\frac{1}{\epsilon} K_+^3\log{K_+})$ operations. The $O(\frac{1}{\epsilon}K_+^3\log{K_+})$ component arises from the add/removal operations, but as we show in Figure \ref{fig:rtcomp}, it is a loose upper bound that scales sub-quadratically in practice. 

\section{Related Work}\label{sec:relwork}
Several proposals have been made for efficient inference with latent feature models. Table \ref{table:wcomp} summarizes the per-iteration complexity of the methods discussed below. In the next section we compare these methods on two synthetic and three real-world datasets.

\citet{velez2009variational} formulated a coordinate ascent variational inference technique for IBP models (VIBP). This method used the ``stick breaking" formulation of the IBP, which maintained coupled beta-distributed priors on the entries of $\IBP$|marginalizing these priors does not allow closed-form MFVB updates. Unlike MEIBP inference, maintaining the beta priors has the undesirable consequence that inactive features contribute to the evidence lower bound and must be ignored when updating the variational distributions. This was not a problem for \citet{velez2009variational}'s finite variational IBP, which computes variational distributions for a linear-Gaussian likelihood with a parametric beta-Bernoulli prior on the latent features.  The inference complexity for both methods is $O(NK_+^2D)$, which is dominated by updating $q(\IBP)$.

\citet{ding2010nonparametric} used mixed expectation-propagation style updates with MFVB inference in order to perform variational inference for a nonnegative linear-Gaussian IBP model (INMF). The expectation-propagation style updates are more complicated than MFVB updates and have per-iteration complexity $O(N(K^3D + KD^2))$. \citet{ding2010nonparametric} motivated this framework by stating that the evidence lower bound of a linear-Gaussian likelihood with a truncated Gaussian prior on the latent factors is negative infinity. This is only true if the variational distribution is a Gaussian, however the free-form variational distribution for their model is a truncated Gaussian, which has a well-defined evidence lower bound.

\citet{doshi2009accelerated} presented a linear-time ``accelerated" Gibbs sampler for conjugate IBP models that effectively marginalized over the latent factors (AIBP).  The per-iteration complexity was $O(N(K^{2} + KD))$. This is comparable to the uncollapsed IBP sampler (UGibbs) that has per-iteration complexity $O(NDK^2)$ but does not marginalize over the latent factors, and as a result, takes longer to mix. In terms of both complexity and empirical performance, the accelerated Gibbs sampler is the most scalable sampling-based IBP inference technique currently available. One constraint of the accelerated IBP is that the latent factor distribution must be conjugate to the likelihood, which for instance, does not allow nonnegative priors on the latent factors.

\citet{rai11ibpbeam} introduced a beam-search heuristic for locating approximate MAP solutions to linear-Gaussian IBP models (BS-IBP). This heuristic sequentially adds a single data point to the model and determines the latent feature assignments by scoring all $2^{K_+}$ latent feature combinations. The scoring heuristic uses an estimate of the joint probability, $P(\Data,\IBP)$ to score assignments, which evaluates the collapsed likelihood $P(\Data|\IBP)$ for all $2^{K_+}$ possible assignments: an expensive $N^3(K_+ + D)$ operation, yielding a per-iteration complexity of $O(N^3(K_+ + D)2^{K_+})$. 

\begin{table}[t]
\caption{Worst-case per-iteration complexity given a linear-Gaussian likelihood model for $N$ $D$-dimensional observations and $K_+$ active latent features.  }
\label{table:wcomp}
\vskip 0.15in
\begin{center}
\begin{small}
\begin{tabular}{|L{3.75cm}|L{3.49cm}|}
\hline
Algorithm & Iteration Complexity\\ \hline
MEIBP  &$O(N(K_+^2D + K_+^3\ln{K_+}))$\\  \hline
VIBP \citep{velez2009variational} &$O(NK_+^2D)$\\  \hline
AIBP \citep{doshi2009accelerated}&$O(N(K_+^{2} + K_+D))$\\ \hline
UGibbs \citep{doshi2009accelerated}&$O(NK_+^{2}D)$\\ \hline
BS-IBP \citep{rai11ibpbeam}&$O(N^3(K_+ +  D)2^{K_+})$\\ \hline
INMF \citep{ding2010nonparametric}&$O(N(K_+^3D + K_+D^2))$ \\ \hline
\end{tabular}
\end{small}
\end{center}
\vskip -0.1in
\end{table}

\section{Experiments}\label{sec:exps}
\begin{figure*}[thb]
	\centering$
	\begin{array}{cc}
	\setlength\fheight{2.75cm} 
	\setlength\fwidth{5.5cm}
	\includegraphics{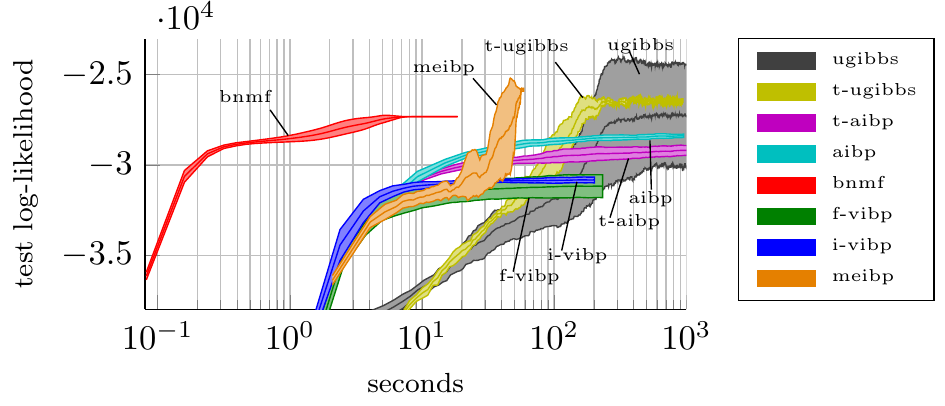}
	& \setlength\fheight{2.75cm} 
	\setlength\fwidth{5.5cm}
	\includegraphics{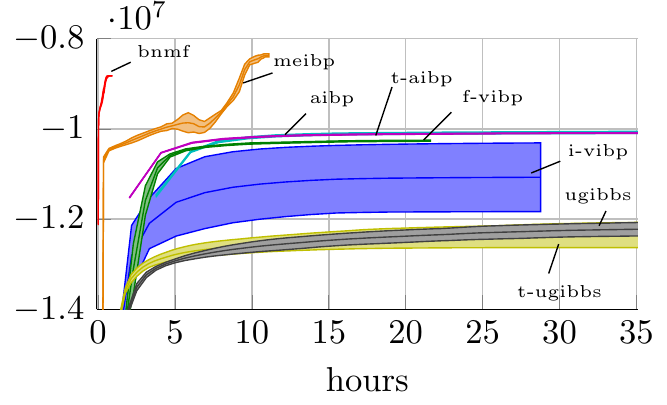}
	\end{array}$
	\caption{Evolution of test log-liklihood over time for a synthetic dataset; \textbf{Left}:  dataset with $N=500, D=500,K=20$ (NB: x-axis is log-scale) \textbf{Right}: dataset with $N=10^5, D=10^3,K=50$ (NB: x-axis is linear-scale). }
	\label{fig:synthetic}
\end{figure*}
We evaluated the inference quality and efficiency of MEIBP inference on two synthetic and three real-world datasets. We used the runtime and predictive likelihood of held-out observations as our performance criteria and compared MEIBP inference with the methods listed in Table \ref{table:wcomp} (the finite and infinite VIBP are differentiated with an ``f-" and ``i-" prefix). We used a truncated Gaussian prior on the latent factors for UGibbs and INMF, and Gaussian priors for the AIBP and variational methods. In our evaluations, we also included \citet{schmidt09ica}'s iterated conditional modes algorithm, which computes a MAP estimate of a parametric nonnegative matrix factorization model: $\Data = \mathbf{B}\Features + \mathbf{E}$, where $\mathbf{B}$ and $\Features$ have exponential priors and $\mathbf{E}$ is zero-mean Gaussian noise. We abbreviate this model ``BNMF";  it has a per-iteration complexity of $O(N(K_+^2 + K_+D))$.

The VIBP and MEIBP inference methods specify a maximum $K$ value, while the sampling methods are unbounded. Therefore, we also included truncated versions of the sampling methods (indicated by a ``t-" prefix) for a fairer comparison. We centered all input data to have a 0-mean for the models with 0-mean Gaussian priors and a 0-minimum for nonnegative models, and all inferred matrices were initialized randomly from their respective priors. Following \citet{doshi2009accelerated}, we fixed the hyperparameters $\sigData$ and $\sigFeat$ to $\frac{3}{4}\sigma$, where $\sigma$ was the standard deviation across all dimensions of the data, and set $\alpha=3$. We ran each algorithm until the multiplicative difference of the average training log-likelihood differed by less than $10^{-4}$ between blocks of five iterations with a maximum runtime of 36 hours. Our experiments used MATLAB implementations of the algorithms, as provided by the respective authors, on 3.20 GHz processors.

\textbf{Synthetic Data} We created high-noise synthetic datasets in the following way: (1) sample $\ibp_{n,k} \sim \text{Bernoulli}(p=0.4)$,  (2) generate $\Features$ with $K$ random, potentially overlapping binary factors, (3) let $\Data = \IBP\Features + \mathbf{E}$, where $ \mathbf{E} \sim \mathcal{N}(0, 1)$. We evaluated the predictive likelihood on $20\%$ of the dimensions from the last half of the data (see supplementary information).

Figure \ref{fig:synthetic} shows the evolution of the test log-likelihood over time for a small dataset with $N=500, D=500, K=20$ and a large dataset with $N=10^5, D=10^3, K=50$. All models were initialized randomly with the true number of latent features, and the error regions display the standard deviation over five random restarts. The BS-IBP and INMF methods were removed from our experiments following the synthetic dataset tests as both methods took at least an order of magnitude longer than the other methods: in 36 hours, the BS-IBP did not complete a single iteration on the small dataset, and the INMF did not complete a single iteration on the large dataset. 

MEIBP converged quickest among the IBP models for both the small and large dataset, while the parametric BNMF model converged much faster than all IBP models. However, the IBP models captured the sparsity of the latent features and the MEIBP and UGibbs eventually outperformed the BNMF on the small dataset, while only the MEIBP outperformed the BNMF on the large dataset. The VIBP methods converged quicker than the sampling counterparts but had trouble escaping local optima. The uncollapsed samplers eventually performed as well as the MEIBP on the small dataset but did not mix to a well-performing distribution for the large dataset.

\begin{figure*}[thb]
\begin{center}$
\begin{array}{ccc}
\setlength\fheight{3.4cm} 
	\setlength\fwidth{3.95cm}
	\includegraphics{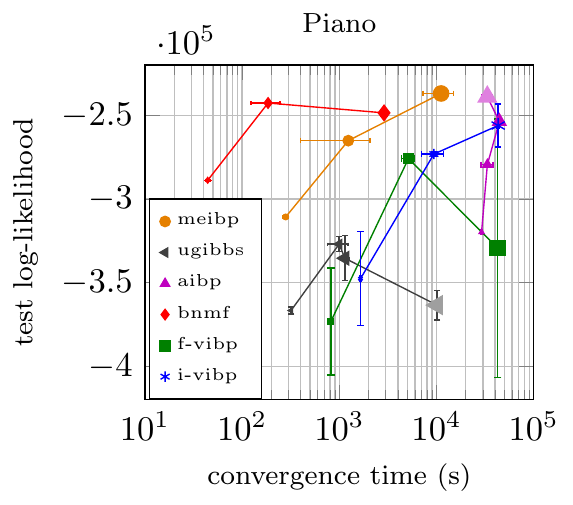} 
	 	&
	\setlength\fheight{3.4cm} 
	\setlength\fwidth{3.95cm}
	\includegraphics{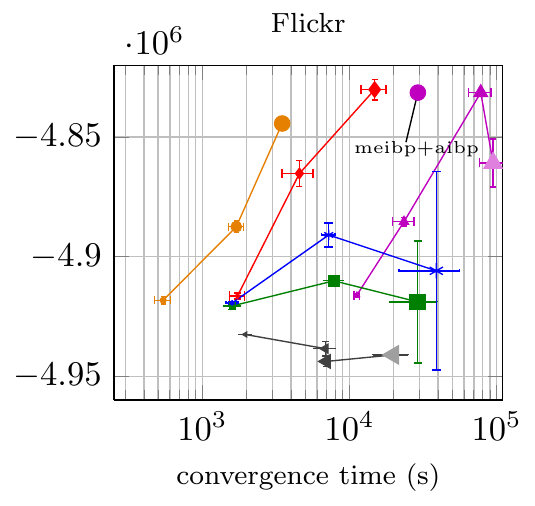} 
	 	&
	\setlength\fheight{3.4cm} 
	\setlength\fwidth{3.95cm}
\includegraphics{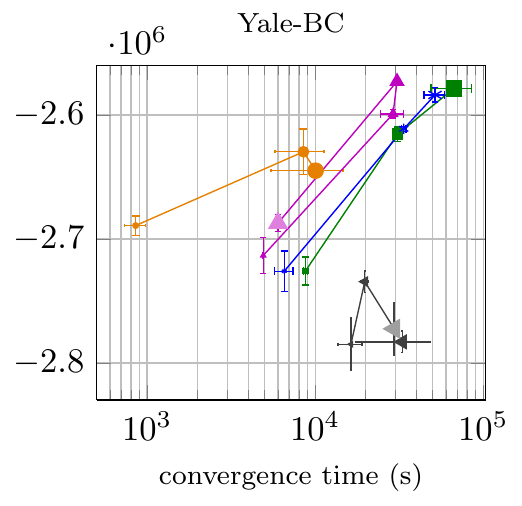} 
\end{array}$
\end{center}
\caption{Inference results on real-world datasets.  The size of the marker indicates the $K$ value for $K=\{10,25,50\}$, with larger markers indicating a larger $K$. AIBP and UGibbs  also include a larger faded marker that shows the inference results for unbounded $K$.  The Flickr plot also shows the result of initializing the AIBP using the MEIBP result. The error-bars indicate the standard deviation of convergence time and test log-likelihood over five random restarts.}
\label{rwresults}
\end{figure*}
\textbf{Real Data} Table \ref{table:rwdata} summarizes the real-world datasets used in our experiments. Piano and Yale-BC are dense real-valued datasets, whereas the Flickr dataset is a sparse binary dataset ($0.81\%$ filled). For the Piano and Flickr datasets, we evaluated the predictive likelihood on a held-out portion of $20\%$ of the dimensions from the last half of the datasets. The Yale-BC dataset had roughly sixty-four facial images of thirty-eight subjects, and we removed the bottom half of five images from each subject for testing.
\begin{table}[thb]
\caption{Summary of real-world datasets. }
\label{table:rwdata}
\vskip 0.15in
\begin{center}
\begin{small}
\begin{tabular}{|L{2.4cm}|l|L{2.4cm}|}
\hline
Dataset & Size ($N\times D$)  &Details\\ \hline
Piano \citep{poliner2006discriminative}  & $16000 \times 161$ & DFT of piano recordings\\  \hline
Flickr \citep{kollar2009utilizing} & $25000 \times  1500$ &binary image-tag indicators\\ \hline
Yale-BC \citep{KCLee05} &$2414 \times 32256$ & face images with various lightings\\  \hline
\end{tabular}\end{small}
\end{center}
\vskip -0.1in
\end{table}

Figure \ref{rwresults} shows the test log-likelihood and convergence time for all inference methods applied to the real-world datasets, averaged over five random restarts. All inference methods were initialized with $K=\{10, 25, 50\}$ as indicated by the size of the marker (the smallest marker shows the $K=10$ results). The sampling methods (AIBP, UGibbs) also include a large faded marker that shows the results for unbounded $K$. 

The Piano results were similar to the small synthetic dataset. The BNMF converged much faster than the IBP models, and the MEIBP performed best among the IBP models in terms of runtime and test log-likelihood|it converged to a similar solution as the AIBP in one-third the time. Though UGibbs has the best per-iteration complexity, it got stuck in poor local optima when randomly initialized. The VIBP methods and MEIBP expressed large uncertainty about the latent factors early on and overcame these poor local optima. By using hard latent feature assignments, the MEIBP took larger steps in the inference space than the VIBP methods, which was beneficial for this dataset, and achieved similar results to the AIBP.

MEIBP inference performed comparable to the best IBP sampling technique for the sparse binary Flickr dataset and converged over an order of magnitude faster. Surprisingly, the dense BNMF inference performed very well on this dataset even though the dataset was sparse and binary. The BNMF converged slower than the MEIBP because it inferred a sparse matrix from a dense prior, which took over four times as many iterations to converge compared to the dense datasets. While the t-AIBP converged to a better solution than the MEIBP, it took over an order-of-magnitude longer to surpass the MEIBP's performance. As we demonstrate with the Flickr results, initializing the AIBP with the MEIBP outcome obtained a similar solution in a fraction of the time (indicated as ``meibp+aibp" on the figure).

The MEIBP converged faster than the other IBP methods for the Yale-BC dataset but to a lower test likelihood. The UGibbs and BNMF also experienced difficulty for this dataset, where BNMF converged to a test log-likelihood around $-3.6 \times 10^{6}$ (not visible in the figure). These linear-Gaussian models with nonnegative priors performed worse than the models with Gaussian priors because the dataset contained many images with dark shadows covering part of the face. The nonnegative priors appeared to struggle with reconstructing these shadows, because unlike the Gaussian priors, they could not infer negative-valued ``shadow" factors that obscured part of the image.

In the above experiments, the MEIBP consistently exhibited a sudden convergence whereby it obtained a local optima and the ls-algorithm did not change any $\IBP$ assignments. This is a characteristic of using hard assignments with a greedy algorithm: at a certain point, changing any latent feature assignments decreased the objective function. This abrupt convergence, in combination with the speed of the ls-algorithm, helped the MEIBP consistently converge faster than other IBP methods. Furthermore, the submodular maximization algorithm converged to local optima that were comparable or better than the sampling or variational results, though at the cost of only obtaining a MAP solution. Like the variational methods, it maintained a distribution over $\Features$ that prevented it from getting stuck in local optima early on, and like the sampling methods, the MEIBP used hard $\IBP$ assignments to take larger steps in the inference space and obtain better optima.

\section{Summary and Future Work}\label{sec:summary}
We presented a new inference technique for IBP models that used \citet{kurihara2008bayesian}'s  ME framework to perform approximate MAP inference via submodular maximization. Our key insight was to exploit the submodularity inherent in the evidence lower bound formulated in $\S$\ref{sec:meibp}, which arose from the quadratic pseudo-Boolean component of the linear-Gaussian model. MEIBP inference converged faster than competing IBP methods and obtained comparable solutions on various datasets. 

There are many discrete Bayesian nonparametric priors, such as the Dirichlet process, and an interesting area for future research will be to generalize our results in order to phrase inference with these priors as submodular optimization problems. Furthermore, we used a simple local-search algorithm to obtain a $\frac{1}{3}$-approximation bound, but concurrently with this work, \citet{buchbinder2012tight} proposed a simpler stochastic algorithm for unconstrained submodular maximization that obtains an expected $\frac{1}{2}$-approximation bound. Using this algorithm, MEIBP inference has an improved worst case complexity of $O(NK_+^2D)$. We will investigate this algorithm in an extended technical version of this paper.

\textbf{Code}: A MATLAB implementation of MEIBP is available at \url{https://github.com/cjrd/MEIBP}.

\textbf{Acknowledgements}: CR was supported by the Winston Churchill Foundation of the United States, and ZG was supported by EPSRC grant EP/I036575/1 and grants from Google and Microsoft. We thank the anonymous reviewers for their helpful comments.

\bibliography{npb_bib}
\bibliographystyle{icml2013stylefiles/icml2013}

\newpage
\appendix
\renewcommand{\thesubsection}{S.\arabic{subsection}}
\section*{Supplementary Material}
\subsection[]{Truncated Gaussian Properties}
In the main text we examined a truncated Gaussian of the form:
\begin{align}
\TN(\tilde{\mu}_{kd}, \tilde{\sigma}_{kd}^2) = 
\frac{2}{\text{erfc}\left(-\frac{\tilde{\mu}_{kd}}{\tilde{\sigma}_{kd}\sqrt{2}}\right)}\mathcal{N}(\tilde{\mu}_{kd}, \tilde{\sigma}_{kd}^2)
\end{align}
with $\mathcal{N}$ representing a Gaussian distribution. The first two moments of $\TN(\tilde{\mu}_{kd}, \tilde{\sigma}_{kd}^2)$ are:
\begin{align}
\mathbb{E}\left[ \feat_{kd} \right] &= \tilde{\mu}_{kd} 
	+  \tilde{\sigma}_{kd}
	\frac{\sqrt{2/\pi}}{\erfcx{\wp_{kd}}}  \label{exa}
	\\
\mathbb{E}\left[ \feat_{kd}^2 \right] &= \tilde{\mu}_{kd}^2 + \tilde{\sigma}_{kd}^2 
	+   \tilde{\sigma}_{kd} \tilde{\mu}_{kd} 
	\frac{\sqrt{2/\pi}}{\erfcx{\wp_{kd}}} \label{exasq}
\end{align}
with $\wp_{kd}=-\frac{\tilde{\mu}_{kd} }{\tilde{\sigma}_{kd}\sqrt{2}}$ and $\erfcx{y}=e^{y^2}(1-\text{erf}(y))$ representing the scaled complementary error function.  The entropy is
\begin{align}
H(q(\feat_{kd})) =&  \frac{1}{2}\ln{\frac{\pi e \tilde{\sigma}_{kd}^2}{2}} 
	+ \ln{\text{erfc}\left(-\frac{\tilde{\mu}_{kd}}{\tilde{\sigma}_{kd} \sqrt{2}} \right)} 
	\\ & + \frac{\tilde{\mu}_{kd}}{\tilde{\sigma}_{kd}} \sqrt{\frac{1}{2\pi}}\left(\text{erfcx}\left(-\frac{\tilde{\mu}_{kd} }{\tilde{\sigma}_{kd}\sqrt{2}}\right)\right)^{-1}.
\end{align}

\subsection{Shifted Equivalence Classes}
\begin{figure*}[htb]
	\centering
	$\vcenter{\hbox{\setlength\fboxsep{0pt}
\setlength\fboxrule{0.5pt}
\fbox{\includegraphics[width=5cm,height=3.5cm]{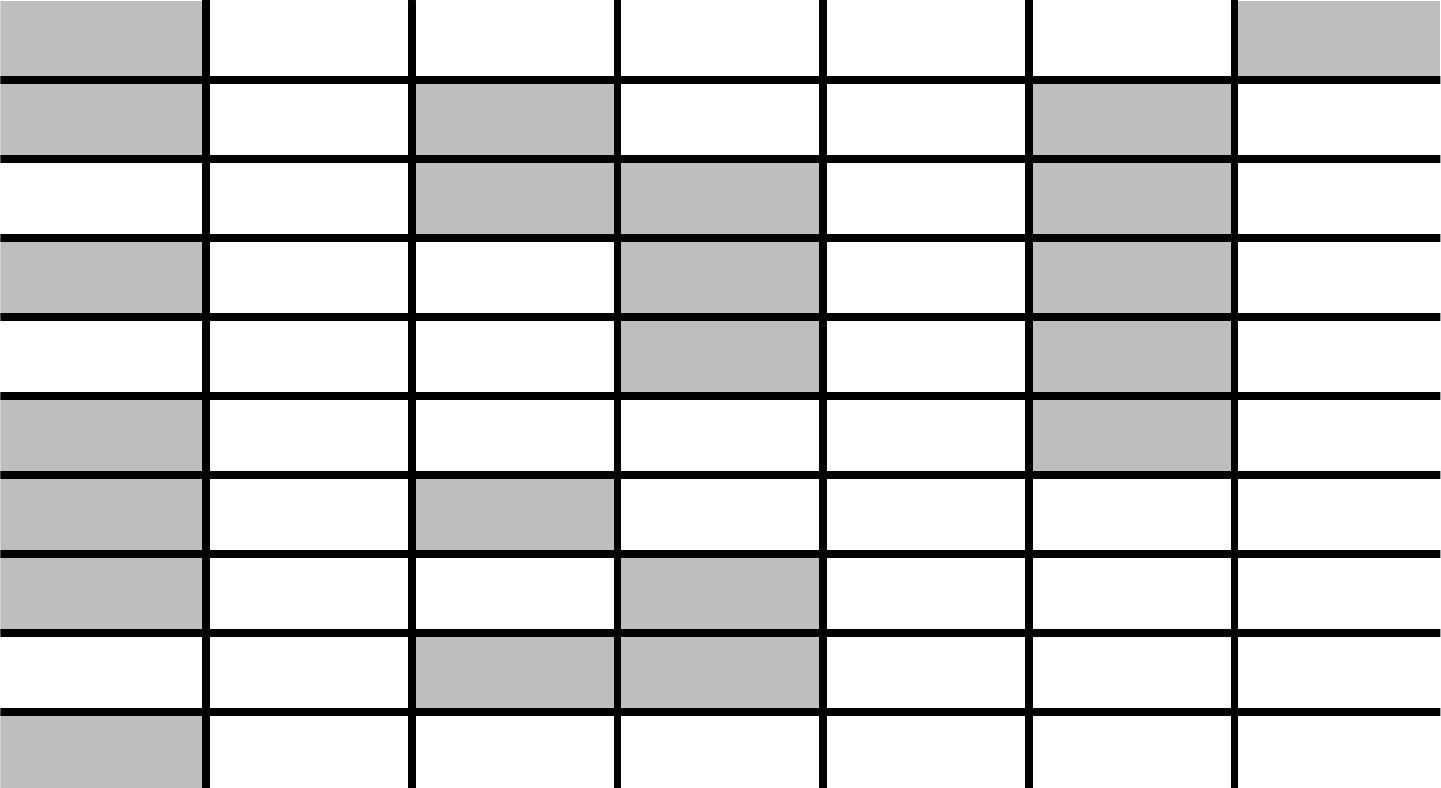}}}}
\vpointer
\vcenter{\hbox{\setlength\fboxsep{0pt}
\setlength\fboxrule{0.5pt}
\fbox{\includegraphics[width=5cm,height=3.5cm]{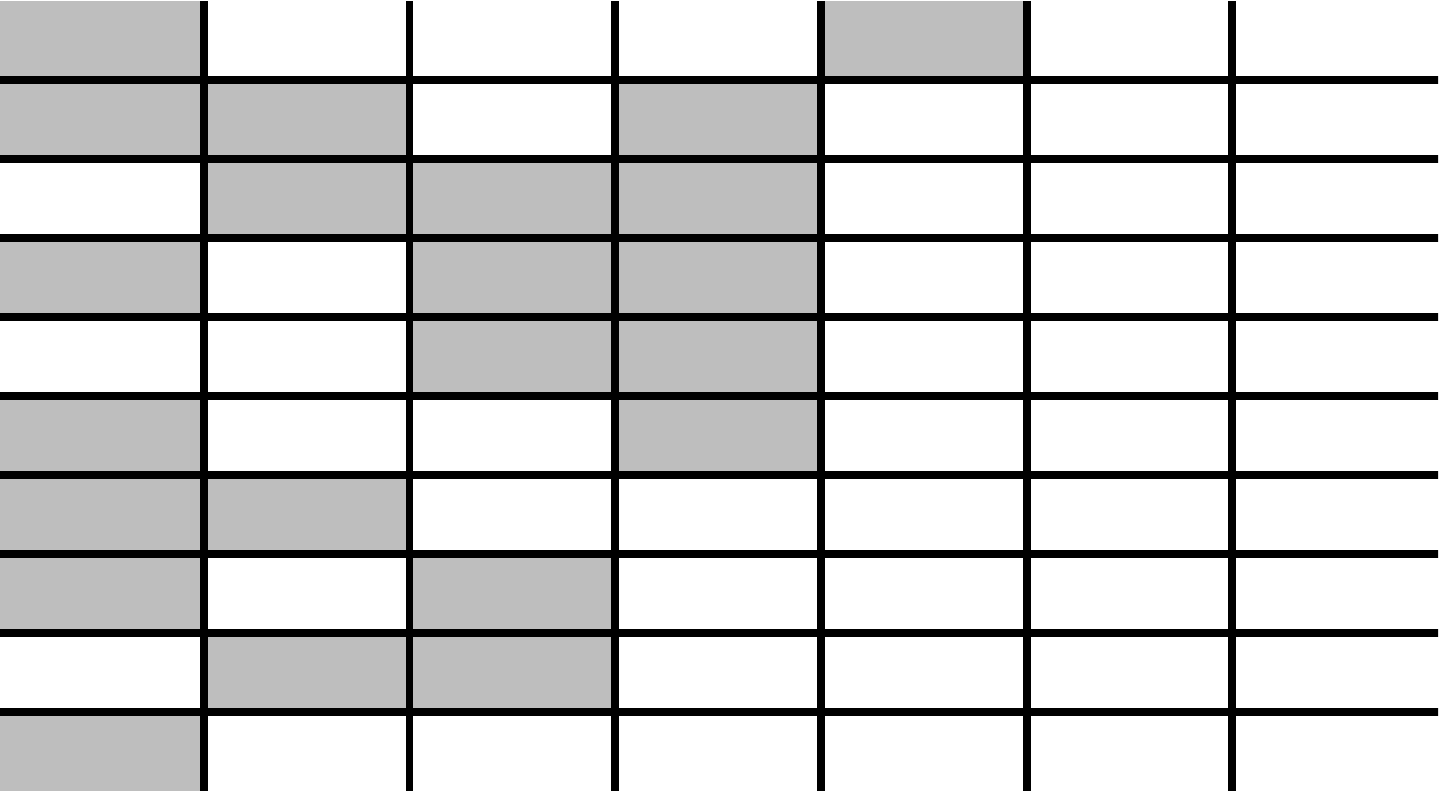}}}}$
	\caption{Example of a binary matrix (left) and its shifted equivalence matrix (dark squares are 1, white squares are 0)|placing the two all-zero columns anywhere in the matrix will yield the same equivalence matrix.}
	\label{fig:sec}
\end{figure*}

Here we discuss the ``shifted" equivalence class of binary matrices first proposed by \citet{ding2010nonparametric}. For a given $N\times K$ binary matrix $\IBP$, the equivalence class for this binary matrix $[ \IBP ]$ is obtained by shifting all-zero columns to the right of the non-zero columns while maintaining the non-zero column orderings, see Figure \ref{fig:sec}. Placing independent $\text{Beta}(\frac{\alpha}{K},1)$ priors on the Bernoulli entries of $\IBP$ and integrating over these priors yields the following probability for $\IBP$, see Eq.\ 27 in \citet{griffiths2005infinite}:
\begin{align}
P(\IBP) = \prod_{k=1}^K\frac{\frac{\alpha}{K}\Gamma(m_k + \frac{\alpha}{K})\Gamma(N-m_k + 1)}{\Gamma(N+1 + \frac{\alpha}{K})}
\end{align}
where $m_k = \sum_{n=1}^N z_{nk}$. Letting $K\rightarrow \infty$ yields $P(\IBP) = 0$ for all $\IBP$. However, the probability of certain equivalence classes of binary matrices, $P([\IBP])$, can remain non-zero as $K \rightarrow \infty$. Specifically, \citet{griffiths2005infinite} show $P([\IBP])$ remains non-zero for the ``left-ordered form" equivalence class of binary matrices, whereby the columns of $\IBP$ are ordered such that the binary values of the columns are non-increasing, where the first row is the most significant bit. Here we outline a similar result for the shifted equivalence class.\footnote{\citet{ding2010nonparametric} proposed this equivalence class but did not explicitly show that it remains well defined as $K\rightarrow \infty$. Furthermore, they did not discuss the collapsed case where we first marginalize over the beta priors on $\IBP$.}

We obtain the probability of the shifted equivalence class by multiplying the multiplicity of the equivalence class by the probability of a matrix within the class. For a given matrix with $K$ columns and $K_+$ non-zero columns, each shifted equivalence class has $\binom{K}{K_+}$ matrices that map to it, yielding:
 \begin{align}
P\left([\IBP]\right) = \binom{K}{K_+}\prod_{k=1}^K\frac{\frac{\alpha}{K}\Gamma(m_k + \frac{\alpha}{K})\Gamma(N-m_k + 1)}{\Gamma(N+1 + \frac{\alpha}{K})}.
\end{align}
Following a similar algebraic rearrangement as \citet{griffiths2005infinite} Eqs.\ 30-33, except replacing the $\frac{K!}{\prod_{h=0}^{2^N-1}K_h!}$ term with $\binom{K}{K_+}$|which occurs because of the different equivalence class multiplicities| results in:
\begin{align}
P([\IBP]) =& \frac{\alpha^{K_+}}{K_+!} \cdot \frac{K!}{(K-K_+)!K^{K_+}} \cdot \left( \frac{N!}{\prod_{j=1}^{2^N-1}(j+\frac{\alpha}{K})}\right)^K  \nonumber
\\ &\cdot \prod_{k=1}^{K_+}\frac{(N-m_k)! \prod_{j=1}^{m_k-1}(j+\frac{\alpha}{K})}{N!}.
\end{align}
We then take the limit $K\rightarrow \infty$ for each of the four terms. The first term has no $K$ dependence and does not change in the infinite limit. For the second term we let $K_0 = K - K+$ and have$\frac{K!}{K_0!K^{K_+}}$. Equations 60-62 in \citet{griffiths2005infinite} show that this term becomes 1 as $K\rightarrow \infty$. The infinite limit of the third and fourth terms are determined in the Appendix of \citet{griffiths2005infinite}. Combining all four terms together yields:
\begin{align}
P([\IBP]) = \frac{\alpha^{K_+}}{K_+!}\text{e}^{-\alpha H_N} \prod_{k=1}^{K^{+}} \frac{(N-m_k)!(m_k - 1)!}{N!}\label{secprior}
\end{align}
where $H_N$ is the $N^{th}$ harmonic number.

The probability of the shifted equivalence class is nearly identical to the probability of the left-ordered-form equivalence class:
\begin{align}\label{lofprior}
P([\IBP]_{\text{lof}}) = \frac{\alpha^{K_+}}{\prod_{h=1}^{2^{N}-1} K_h}\text{e}^{-\alpha H_N} \prod_{k=1}^{K^{+}} \frac{(N-m_k)!(m_k - 1)!}{N!},
\end{align}
where $K_h$ is the number of columns of $\IBP$ with binary value $h\in \{1, \ldots, 2^{N-1}\}$ when the first row is taken to be the most significant bit. The only difference between Eq.\ \ref{secprior} and Eq.\ \ref{lofprior} is the denominator of the first fraction. For the left-ordered-form, this term penalizes $\IBP$ matrices with identical columns. In the feature assignment view, this term penalizes features that are assigned to the exact same set of observations. The $K_+!$ term in the shifted equivalence class prior does not distinguish between identical and distinct columns of $\IBP$, and in turn, does not penalize repeated feature assignments. These two equivalence class probabilities are proportional in the limit of large $N$ as the probability of two columns being identical approaches 0.

\subsection{Hyperparameter Inference}
In the main text we assumed the hyperparameters $\boldsymbol{\theta} = \{\sigData,\sigFeat, \alpha\}$ were known (i.e.\ estimated from the data). Placing conjugate gamma hyperpriors on these parameters allows for a straightforward extension in which we infer their values. Formally, let 
\begin{align}
p(\tau_X) &= \text{Gamma}(\tau_X; a_X, b_X) \\
p(\tau_A) &=\text{Gamma}(\tau_A; a_A, b_A) \\
p(\alpha) &= \text{Gamma}(\alpha; a_\alpha, b_\alpha) 
\end{align}
where $\tau$ represents the precision, equivalent to the inverse variance $\frac{1}{\sigma^2}$, for the variance parameter indicated in the subscript. Update equations for the variational distributions follow from standard update equations for variational inference in exponential families, cf.\ \citet{attias2000variational}, and yield:
\begin{align}
q(\tau_X) &= \text{Gamma}(\widetilde{a}_X, \widetilde{b}_X) \\
q(\tau_A) &= \text{Gamma}(\widetilde{a}_A, \widetilde{b}_A) \\
q(\alpha) &= \text{Gamma}(\widetilde{a}_\alpha, \widetilde{b}_\alpha)
\end{align}
with variance updates
\begin{align}
\widetilde{a}_A &= a_A + \frac{KD}{2}\\
\widetilde{b}_A &= b_A + \frac{1}{2}\sum_{k=1}^{K_+}\sum_{d=1}^D \mathbb{E}\left[ a_{kd}^2 \right]
\end{align}
and
\begin{align}
\widetilde{a}_X &= a_X + \frac{ND}{2}\\
\widetilde{b}_X &= b_X + \frac{1}{2}\sum_{n=1}^N\sum_{d=1}^D\Bigl[ x_{nd}^2 + \sum_{k=1}^{K_+}\Bigl[ \mathbb{E}\left[ a_{kd}^2 \right]z_{nk} \\
& - 2\mathbb{E}[a_{kd}]z_{nk}x_{nd} + 2\sum_{k^{'}=k+1}^{K_+}z_{nk}z_{nk^{'}}a_{kd}a_{k^{'}d}\Bigr] \Bigr]
\end{align}
and $q(\alpha)$ updates
\begin{align}
\widetilde{a}_\alpha &= a_\alpha + K_+\\
\widetilde{b}_\alpha &= b_\alpha + H_N.
\end{align}
MEIBP inference is carried out exactly as discussed in the main text except all instances of $\sigData,\sigFeat$, and $\alpha$ are replaced with the expectation from their respective variational distribution. Furthermore the variational lower bound also has three additional entropy terms for gamma distributions, one for each hyperparameter.

\subsection{Evidence as a function of $\mathbf{Z}_{n\cdot}$}
As shown in the main text, we obtain a submodular objective function for each $\IBP\lcdot{n}$, $n\in\{1,\ldots,N\}$ by examining the evidence as a function of $\IBP\lcdot{n}$ while holding constant all $n'\in \{1,\ldots,N\}\setminus n$. The evidence is
\begin{align}
&\frac{1}{\sigData^2}\sum_{n=1}^N\left[ -\frac{1}{2}\IBP\lcdot{n}\ExpFM\ExpFM^{T}\IBP\lcdot{n}^{T} + \IBP\lcdot{n}\boldsymbol{\xi}\lcdot{n}^T\right] - \ln{K_+!} \nonumber \\
 &~~~~+ \sum_{k=1}^{K^+}\left[\ln{\frac{(N-m_k)!(m_k - 1)!}{N!}}
 + \eta_k \right]
+  \text{const} \label{vlbreform}\\
&\xi_{nk} = \ExpFM\lcdot{k}\Data\lcdot{n}^T + \frac{1}{2}\sum_{d=1}^D\left[\E[\feat_{kd}]^2 - \E[\feat_{kd}^2] \right] \label{xiterm}\\
&\eta_k = \sum_{d=1}^D\left[ - \frac{\ln{\frac{\pi \sigFeat^2}{2\alpha^{2/D}}}}{2} - \frac{\E[a_{kd}^2]}{2\sigFeat^2} + H(q(\feat_{kd})) \right],
\end{align}
which nearly factorizes over the $\IBP_{n\cdot}$ because the likelihood component and parts of the prior components naturally fit into a quadratic function of $\IBP_{n\cdot}$. The $\ln K_+!$ and $\eta_k$ only couple the rows of $\IBP$ when $K_+$ changes, while the log-factorial term couples the rows of $\IBP$ through the sums of the columns. Both of these terms only depend on statistics of $\IBP$ (the $m_k$ values and $K_+$), not the $\IBP$ matrix itself, e.g.\ permuting the rows of $\IBP$ would not affect these terms. Furthermore, $\ln{K_+}$ and $\eta_k$ have no $N$ dependence and become insignificant as $N$ increases. These observations, in conjunction with the MEIBP performance in the experimental section of the main text, indicate that optimizing Eq.\ \ref{vlbreform} for $\IBP_{n\cdot}$ is a reasonable surrogate for optimizing $\IBP$.

Here we explicitly decompose Eq.\ \ref{vlbreform} to show its $\IBP_{n\cdot}$ dependency. Decomposing $\ln{\frac{(N-m_k)!(m_k - 1)!}{N!}}$ is straightforward if we first define the function:
\begin{equation} 
  \nu(\ibp_{nk}) = \begin{cases}
    \ln{(N-m_{k\setminus n} - \ibp_{nk})!(m_{k \setminus n} + \ibp_{nk} - 1)!/N!} \label{nueq}\\ 
    0,  ~~\text{if $m_{k\setminus n} = 0$ and $\ibp_{nk} = 0$}.
  \end{cases}
\end{equation}
where the ``$\setminus n$" subscript indicates the variable with the $n^{\text{th}}$ row removed from $\IBP$. For a given $n$ we have:
\begin{align*}
\sum_{k=1}^{K_+} \nu(\ibp_{nk}) = & \sum_{k=1}^{K_+}\ln{(N-m_k)!(m_k - 1)!/N!}  \\
=& \sum_{k=1}^{K_+} \ibp_{nk}\left( \nu(\ibp_{nk}=1) - \nu(\ibp_{nk}=0) \right) \\ &+ \nu(\ibp_{nk}=0) \numberthis,
\end{align*}
which makes the $\IBP\lcdot{n}$ dependency explicit and lets us add $\nu(\ibp_{nk}=1) - \nu(\ibp_{nk}=0)$  into the inner-product term, $\boldsymbol{\xi}\lcdot{n}$, and place $\nu(\ibp_{nk}=0) $ into a constant term.  We can incorporate $\eta_k$ into the inner-product term in a similar manner for a given $n \in \{1,\ldots,N\}$ :
\begin{align}
\sum_{k=1}^{K_+}\eta_k &= \sum_{k:m_{k\setminus n} > 0} \eta_k  + \sum_{k=1}^{K_+}\boldsymbol{1}_{\{m_{k\setminus n} = 0\}}\ibp_{nk}\eta_k,
\end{align}
where the first term does not depend on $\IBP\lcdot{n}$ and is added to the constant term, while the second term is added to the inner-product term. 
Finally, for a given $n \in \{1,\ldots,N\}$ the $\ln{K!}$ term becomes
\begin{align}
\ln{K_+!} =  \ln{\left(K_{+\setminus n} + \sum_{k=1}^{K_+}\left[ \boldsymbol{1}_{\{m_{k\setminus n} = 0\}}\ibp_{nk}\right] \right)!},
\end{align}
where $\boldsymbol{1}_{\{ \cdot \}}$ is the indicator function. As stated in the main text, combining the above terms yields the following submodular objective function for $n=1,\ldots,N$:
\begin{align}
 \mathcal{F}(\IBP\lcdot{n}) =& -\frac{1}{2\sigData^2}\IBP\lcdot{n}\ExpFM\ExpFM^{T}\IBP\lcdot{n}^{T} + \IBP\lcdot{n}\OptMat\lcdot{n}^{T} + \textit{const} \nonumber\\
  &-  \ln{ \left(K_{+\setminus n} + \sum_{k=1}^{K_+}\left[ \boldsymbol{1}_{\{m_{k\setminus n} = 0\}}\ibp_{nk}\right] \right)!} \label{subcf2}\\
\ExpFM\lcdot{k} =& \left(\E\left[ a_{k1}\right], \ldots,  \E\left[ a_{kD}\right]  \right)\\
\omega_{nk} =& \frac{1}{\sigData^2}\left(\ExpFM\lcdot{k}\Data\lcdot{n}^T + \frac{1}{2}\sum_{d=1}^D\left[\E[\feat_{kd}]^2 - \E[\feat_{kd}^2] \right]\right) \nonumber \\
& +  \nu(\ibp_{nk}=1) - \nu(\ibp_{nk}=0)+ \boldsymbol{1}_{\{m_{k\setminus n} = 0\}}\eta_k,
\end{align}
$\boldsymbol{1}\{\cdot\}$ is the indicator function, and the subscript $``\setminus n"$ is the value of the given variable after removing the $n^\textit{th}$ row from $\IBP$.

\subsection{Additional MEIBP Characterization}
In this section, we will maintain a growing list of additional MEIBP characterization experiments. See \url{http://arxiv.org/abs/1304.3285} for the current version.
\subsubsection{Learning $K_+$}\label{sec:learningK}
An ostensible advantage of using Bayesian nonparametric priors is that a user does not need to specify the multiplicity of the prior parameters. Clever sampling techniques such as slice sampling and retrospective sampling allow samples to be drawn from these nonparametric priors, c.f.\ \citet{teh2007stick} and \citet{papaspiliopoulos2008retrospective}. However variational methods are not directly amenable to Bayesian nonparametric priors as the variational optimization cannot be performed over an unbounded prior space. Instead, variational methods must specify a maximum model complexity (parameter multiplicity). Several heuristics have been proposed to address this limitation: \citet{wang2012truncation} sampled from the variational distribution for the local parameters|which included sampling from the unbounded prior| and used the empirical distributions of the local samples to update the global parameters, while \citet{ding2010nonparametric} simply started with $K_+ =1$ and greedily added features. We did not address these techniques in this work as the MEIBP performed competitively with the unbounded sampling techniques without employing these types of heuristics. Furthermore, here we demonstrate that the MEIBP can robustly infer the true number of latent features when the $K_+$ bound is greater than the true number of latent features.

\begin{figure}[thb]
\center
\includegraphics[height=1.5cm,width=7cm]{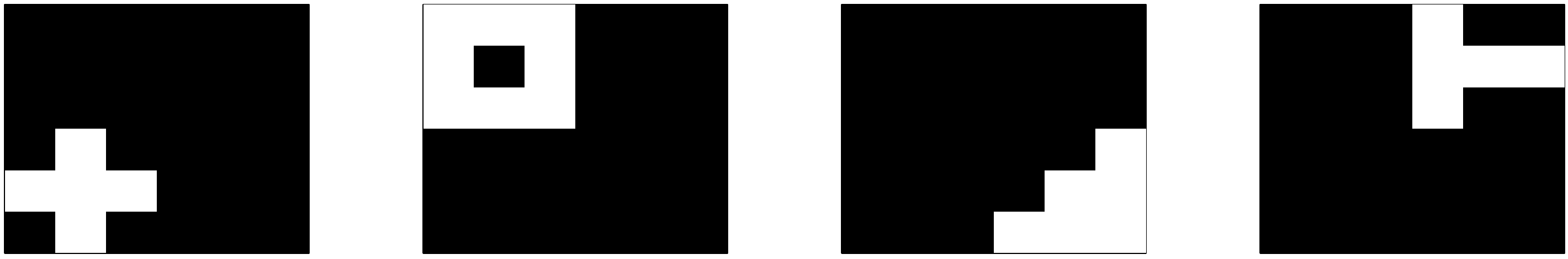}
\caption{The four binary latent factors used in the sensitivity analysis in this section. The white squares are ones and the dark squares are zeros.}
\label{fig:binfacts}
\end{figure}

\begin{figure}[thb]
\begin{center}$
\begin{array}{c}
	\includegraphics[height=1.5cm,width=7cm]{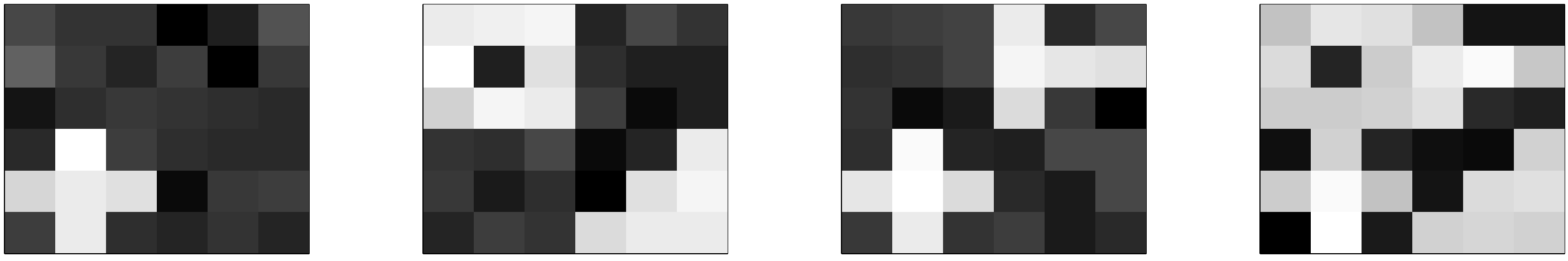} 
	 	\\
	\includegraphics[height=1.5cm,width=7cm]{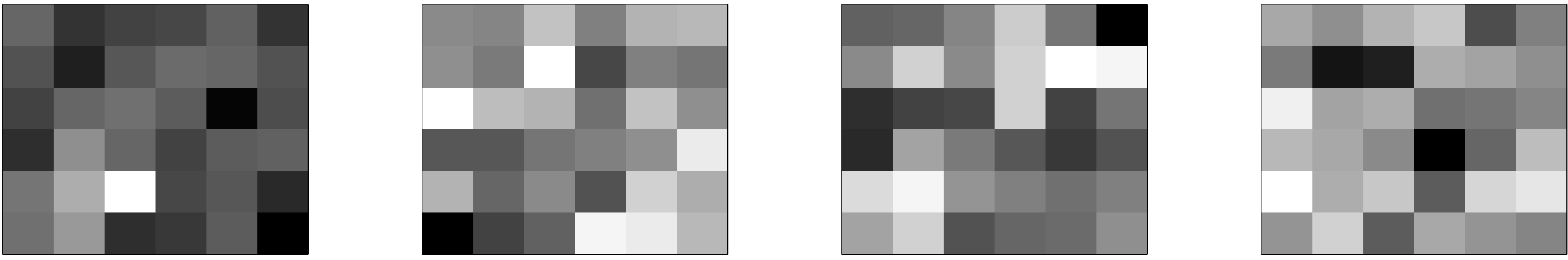} 
	 	\\
		\includegraphics[height=1.5cm,width=7cm]{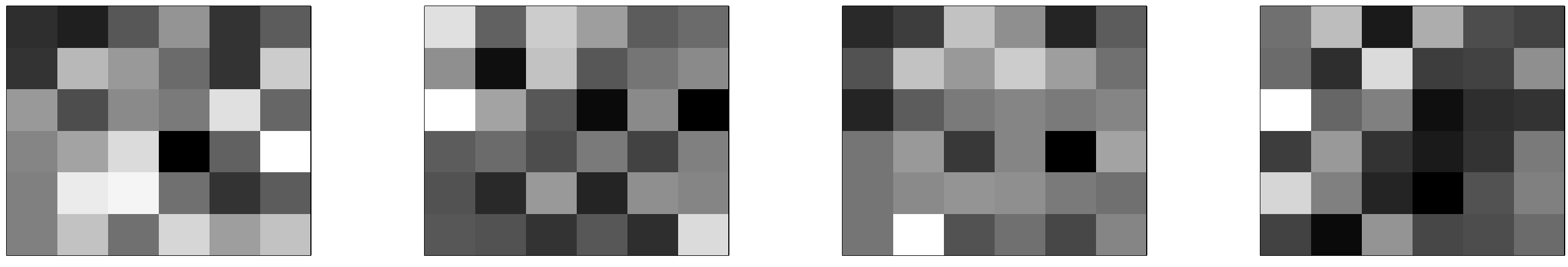} 
\end{array}$
\end{center}
\caption{Example data used in the sensitivity analysis discussed in $\S$\ref{sec:learningK}. Each column contains the same combination of latent factors, where the top row has a data noise term of $\sigma_X = 0.1$, the middle row has $\sigma_X = 0.5$, and the bottom row has $\sigma_X = 1.0$. Top: histogram of final $K_+$ value. Bottom: final $K_+$ value as a function of $\sigma_X$.}
\label{fig:noisydata}
\end{figure}

For this experiment we generated the binary images dataset used in \citet{griffiths2005infinite}, where the dataset, $\Data$, consisted of $2000$ $6\times 6$ images. Each row of $\Data$ was a $36$ dimensional vector of pixel intensity values that was generated by using $\IBP$ to  linearly combine a subset of the four binary factors shown in Figure \ref{fig:binfacts}. Gaussian white noise, $\mathcal{N}(0,\sigma_X)$, was then added to each image, yielding $\Data = \IBP \Features + \mathbf{E}$. The feature vectors, $\IBP\lcdot{n}$ were sampled from a distribution in which each factor was present with probability $0.5$. Figure \ref{fig:noisydata} shows four of these images with different $\sigma_X$ values.

\begin{figure}[thb]
\begin{center}$
\begin{array}{c}
\setlength\fheight{2.8cm} 
	\setlength\fwidth{5.95cm}
%
%
%
%
\begin{tikzpicture}

\begin{axis}[%
width=\fwidth,
height=\fheight,
scale only axis,
xmin=1, xmax=8,
xtick={1,2,3,4,5,6,7,8,9,10},
xlabel={$K_+$},
ymin=0, ymax=70,
ylabel={count}]
\definecolor{mycolor12}{rgb}{0.9,0.5,0}

\addplot [fill=mycolor12,draw=black,forget plot] coordinates{ (0.5,0)(0.5,0)(1.5,0)(1.5,0)};
\addplot [fill=mycolor12,draw=black,forget plot] coordinates{ (1.5,0)(1.5,1)(2.5,1)(2.5,0)};
\addplot [fill=mycolor12,draw=black,forget plot] coordinates{ (2.5,0)(2.5,19)(3.5,19)(3.5,0)};
\addplot [fill=mycolor12,draw=black,forget plot] coordinates{ (3.5,0)(3.5,68)(4.5,68)(4.5,0)};
\addplot [fill=mycolor12,draw=black,forget plot] coordinates{ (4.5,0)(4.5,8)(5.5,8)(5.5,0)};
\addplot [fill=mycolor12,draw=black,forget plot] coordinates{ (5.5,0)(5.5,1)(6.5,1)(6.5,0)};
\addplot [fill=mycolor12,draw=black,forget plot] coordinates{ (6.5,0)(6.5,3)(7.5,3)(7.5,0)};
\addplot [fill=mycolor12,draw=black,forget plot] coordinates{ (7.5,0)(7.5,0)(8.5,0)(8.5,0)};
\addplot [fill=mycolor12,draw=black,forget plot] coordinates{ (8.5,0)(8.5,0)(9.5,0)(9.5,0)};
\addplot [fill=mycolor12,draw=black,forget plot] coordinates{ (9.5,0)(9.5,0)(10.5,0)(10.5,0)};
\end{axis}
\end{tikzpicture}%
	 	\\
	\setlength\fheight{2.8cm} 
	\setlength\fwidth{5.95cm}
%
%
%
%
\begin{tikzpicture}
\definecolor{mycolor12}{rgb}{0.9,0.5,0}
\begin{axis}[%
width=\fwidth,
height=\fheight,
scale only axis,
xmin=0, xmax=1.01,
xlabel={$\sigma{}_\text{X}$},
xmajorgrids,
ymin=1.5, ymax=7.5,
ylabel={$K_+$},
ytick={2,3,4,5,6,7},
ymajorgrids]
\addplot [
color=mycolor12,
mark size=2.0pt,
only marks,
mark=*,
mark options={solid},
forget plot
]
coordinates{
 (0.1,4)(0.109090909090909,4)(0.118181818181818,4)(0.127272727272727,4)(0.136363636363636,4)(0.145454545454545,4)(0.154545454545455,4)(0.163636363636364,4)(0.172727272727273,3)(0.181818181818182,4)(0.190909090909091,4)(0.2,3)(0.209090909090909,3)(0.218181818181818,4)(0.227272727272727,3)(0.236363636363636,3)(0.245454545454545,3)(0.254545454545455,4)(0.263636363636364,4)(0.272727272727273,4)(0.281818181818182,5)(0.290909090909091,4)(0.3,4)(0.309090909090909,4)(0.318181818181818,3)(0.327272727272727,3)(0.336363636363636,3)(0.345454545454545,3)(0.354545454545454,4)(0.363636363636364,4)(0.372727272727273,3)(0.381818181818182,4)(0.390909090909091,4)(0.4,4)(0.409090909090909,4)(0.418181818181818,4)(0.427272727272727,5)(0.436363636363636,4)(0.445454545454545,4)(0.454545454545455,4)(0.463636363636364,3)(0.472727272727273,4)(0.481818181818182,4)(0.490909090909091,4)(0.5,4)(0.509090909090909,4)(0.518181818181818,3)(0.527272727272727,4)(0.536363636363636,3)(0.545454545454546,4)(0.554545454545455,4)(0.563636363636364,4)(0.572727272727273,4)(0.581818181818182,3)(0.590909090909091,4)(0.6,3)(0.609090909090909,4)(0.618181818181818,4)(0.627272727272727,2)(0.636363636363636,4)(0.645454545454545,3)(0.654545454545455,4)(0.663636363636364,3)(0.672727272727273,4)(0.681818181818182,4)(0.690909090909091,4)(0.7,4)(0.709090909090909,4)(0.718181818181818,3)(0.727272727272727,4)(0.736363636363636,4)(0.745454545454545,4)(0.754545454545455,4)(0.763636363636364,4)(0.772727272727273,4)(0.781818181818182,4)(0.790909090909091,5)(0.8,4)(0.809090909090909,4)(0.818181818181818,4)(0.827272727272727,4)(0.836363636363636,4)(0.845454545454545,4)(0.854545454545454,4)(0.863636363636364,4)(0.872727272727273,4)(0.881818181818182,4)(0.890909090909091,5)(0.9,4)(0.909090909090909,5)(0.918181818181818,4)(0.927272727272727,4)(0.936363636363636,5)(0.945454545454545,5)(0.954545454545455,7)(0.963636363636364,5)(0.972727272727273,7)(0.981818181818182,4)(0.990909090909091,6)(1,7) 
};
\end{axis}
\end{tikzpicture}%
\end{array}$
\end{center}
\caption{Final feature count ($K_+$ value) for MEIBP inference where $K_{\text{true}} = 4$ for the binary image data with $K_+$ initialized to $20$ for $\sigma_X = 0.1,\ldots,1.0$ in 100 evenly spaced increments with all hyperparameters  and algorithm options fixed during the experiment. }
\label{fig:k-var-meibp}
\end{figure}
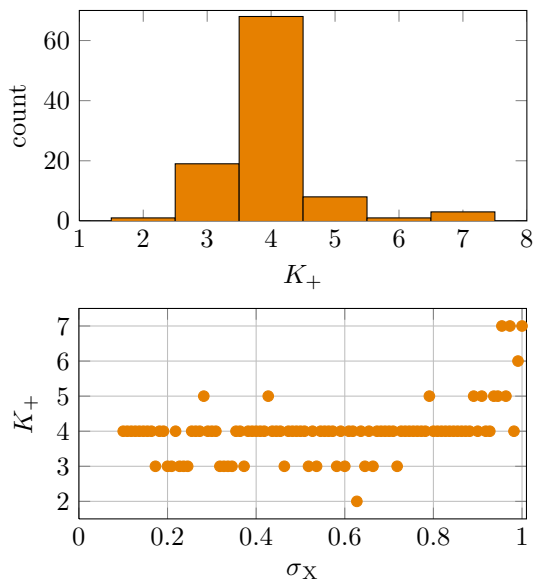

We initialized the MEIBP with $K=20$, $\sigma_X$=1.0, $\sigma_A = 1.0$, $\alpha = 2$, $\tilde{\mu}_{kd} \sim |\mathcal{N}(0, 0.05)|$ (variational factor means),  $\tilde{\sigma}_{kd} \sim |\mathcal{N}(0, 0.1)|$ (variational factor standard deviations), $z_{nk} \sim \text{Bernoulli}(\frac{1}{3})$. With this initialization, we tested the MEIBP robustness by performing MEIBP inference on $\Data$  for $\sigma_X = 0.1,\ldots,1.0$ in 100 evenly spaced increments with all hyperparameters  and algorithm options unchanged during the experiment. MEIBP convergence was determined in the same way as the main experimental section. Figure \ref{fig:k-var-meibp} (top) shows a histogram of the final number of MEIBP features ($K_{\text{true}} = 4$) and Figure \ref{fig:k-var-meibp} (bottom) shows the final number of MEIBP features as a function of $\sigma_X$.

These results indicate that the regularizing nature of the IBP prior tends to lead to the correct number of latent features even when the $K_+$ bound is much larger than the true $K_+$. Furthermore this experiment indicates that MEIBP inference is robust to model noise, at least, for the simple data used in this experiment. At a medium level of data noise, the inference occasionally finished with $K_+ = 3$, which resulted from two true latent factors collapsing to the same inferred latent feature. Once this occurred, MEIBP did not have a mechanism for splitting the features. For $\sigma_X$ comparable to the latent factors, $\sigma_X \geq 0.9$, MEIBP often inferred ``noise features," which were essentially whitenoise and were typically active for less than $4\%$ of the data instances. In future experiments we will attempt to flesh out the practical differences between unbounded priors and priors that operate in a large bounded latent space.

\end{document}